\documentclass[10pt,twocolumn,letterpaper]{article}

\usepackage{cvpr}
\usepackage{times}
\usepackage{booktabs}
\usepackage{epsfig}
\usepackage{graphicx}
\usepackage{amsmath}
\usepackage{amssymb}
\usepackage{bm}
\usepackage{subfigure}
\usepackage{multirow}
\usepackage{array}
\usepackage{enumitem}
\usepackage{enumerate}
\usepackage{mathrsfs}
\usepackage{amsmath}
\usepackage{rotating}
\usepackage{amssymb}
\usepackage[utf8]{inputenc}
\usepackage[english]{babel}
\usepackage[numbers,sort]{natbib}

\usepackage[breaklinks=true,bookmarks=false]{hyperref}

\cvprfinalcopy 

\def\cvprPaperID{5105} 

\begin{document}

\title{BiDet: An Efficient Binarized Object Detector}

\author{Ziwei Wang\textsuperscript{1,2,3},
    Ziyi Wu\textsuperscript{1},
	Jiwen Lu\textsuperscript{1,2,3,}\thanks{Corresponding author},
	Jie Zhou\textsuperscript{1,2,3,4}\\	
\textsuperscript{1} Department of Automation, Tsinghua University, China\\
\textsuperscript{2} State Key Lab of Intelligent Technologies and Systems, China\\
\textsuperscript{3} Beijing National Research Center for Information Science and Technology, China\\
\textsuperscript{4} Tsinghua Shenzhen International Graduate School, Tsinghua University, China\\
	{\tt \small \{wang-zw18, wuzy17\}@mails.tsinghua.edu.cn; \{lujiwen,jzhou\}@tsinghua.edu.cn}
}

\maketitle
\begin{abstract}
In this paper, we propose a binarized neural network learning method called BiDet for efficient object detection. Conventional network binarization methods directly quantize the weights and activations in one-stage or two-stage detectors with constrained representational capacity, so that the information redundancy in the networks causes numerous false positives and degrades the performance significantly. On the contrary, our BiDet fully utilizes the representational capacity of the binary neural networks for object detection by redundancy removal, through which the detection precision is enhanced with alleviated false positives. Specifically, we generalize the information bottleneck (IB) principle to object detection, where the amount of information in the high-level feature maps is constrained and the mutual information between the feature maps and object detection is maximized. Meanwhile, we learn sparse object priors so that the posteriors are concentrated on informative detection prediction with false positive elimination. Extensive experiments on the PASCAL VOC and COCO datasets show that our method outperforms the state-of-the-art binary neural networks by a sizable margin.\footnote{Code: \href{https://github.com/ZiweiWangTHU/BiDet.git}{https://github.com/ZiweiWangTHU/BiDet.git}}
\end{abstract}

\begin{figure}[t]
	\centering
	\includegraphics[height=6cm, width=8cm]{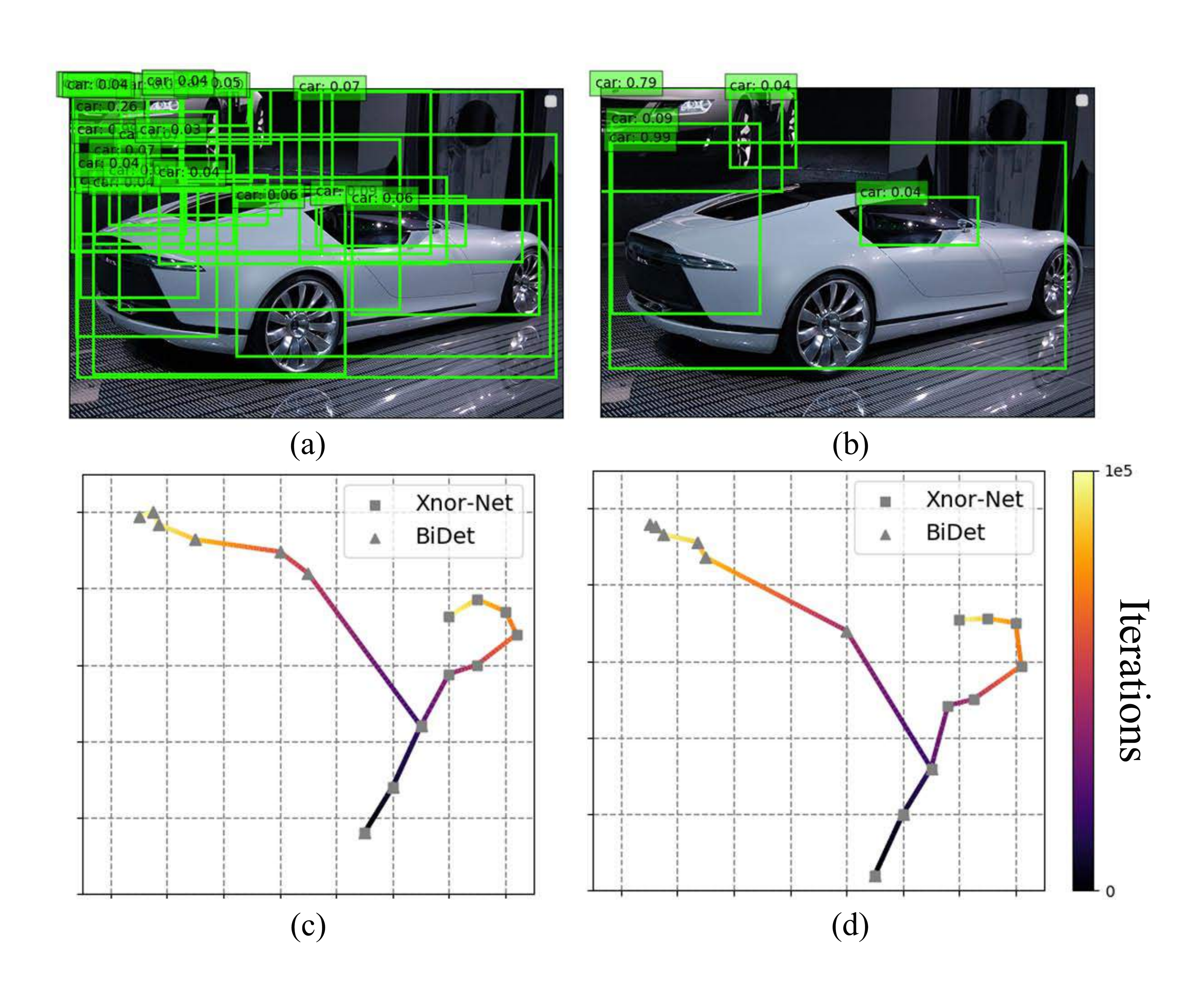}
	\caption{An example of the predicted objects with the binarized SSD detector on PASCAL VOC. (a) and (b) demonstrate the detection results via Xnor-Net and the proposed BiDet, where the false positives are significantly reduced in our method. (c) and (d) reveal the information plane dynamics for the training set and test set respectively, where the horizontal axis means the mutual information between the high-level feature map and input and the vertical axis represents the mutual information between the object and the feature map. Compared with Xnor-Net, our method removes the redundant information and fully utilizes the network capacity to achieve higher performance. (best viewed in color).}
	\vspace{-0.5cm}
	\label{fp}
\end{figure}

\section{Introduction}
Convolutional neural network (CNN) based object detectors \cite{girshick2015fast,he2017mask,ren2015faster,liu2016ssd,lin2017focal} have achieved state-of-the-art performance due to the strong discriminative power and generalization ability. However, the CNN based detection methods require massive computation and storage resources to achieve ideal performance, which limits their deployment on mobile devices. Therefore, it is desirable to develop detectors with lightweight architectures and few parameters.

To reduce the complexity of deep neural networks, several model compression methods have been proposed including pruning \cite{molchanov2019importance,zhao2019variational,he2017channel}, low-rank decomposition \cite{lin2018holistic,peng2018extreme,kim2019efficient}, quantization \cite{wang2019learning,li2019fully,gong2019differentiable}, knowledge distillation \cite{wei2018quantization,wang2019private,chen2017learning}, architecture design \cite{sandler2018mobilenetv2,zhang2018shufflenet,qin2019thundernet} and architecture search \cite{wu2019fbnet,tan2019mnasnet}. Among these methods, network quantization reduces the bitwidth of the network parameters and activations for efficient inference. In the extreme cases, binarizing weights and activations of neural networks decreases the storage and computation cost by $32\times$ and $64\times$ respectively. However, deploying binary neural networks with constrained representational capacity in object detection causes numerous false positives due to the information redundancy in the networks.

In this paper, we present a BiDet method to learn binarized neural networks including the backbone part and the detection part for efficient object detection. Unlike existing methods which directly binarize the weights and activations in one-stage or two-stage detectors, our method fully utilizes the representational capacity of the binary neural networks for object detection via redundancy removal, so that the detection precision is enhanced with false positive elimination. More specifically, we impose the information bottleneck (IB) principle on binarized object detector learning, where we simultaneously limit the amount of information in high-level feature maps and maximize the mutual information between object detection and the learned feature maps. Meanwhile, the learned sparse object priors are utilized in IB, so that the posteriors are enforced to be concentrated on informative prediction and the uninformative false positives are eliminated. Figure \ref{fp} (a) and (b) show an example of predicted positives obtained by Xnor-Net \cite{rastegari2016xnor} and our BiDet respectively, where the false positives are significantly reduced in the latter. Figure \ref{fp} (c) and (d) depict the information plane dynamics for the training and test sets respectively, where our BiDet removes the information redundancy and fully utilizes the representational power of the networks. Extensive experiments on the PASCAL VOC \cite{everingham2010pascal} and COCO \cite{lin2014microsoft} datasets show that our BiDet outperforms the state-of-the-art binary neural networks in object detection across various architectures. Moreover, BiDet can be integrated with other compact object detectors to acquire faster speedup and less storage. Our contributions include:
\begin{itemize}[leftmargin=*]
	\item To the best of our knowledge, we propose the first binarized networks containing the backbone and detection parts for efficient object detection.
	\item We employ the IB principle for redundancy removal to fully utilize the capacity of binary neural networks and learn the sparse object priors to concentrate posteriors on informative detection prediction, so that the detection accuracy is enhanced with false positive elimination.
	\item We evaluate the proposed BiDet on the PASCAL VOC and the large scale COCO datasets for comprehensive comparison with state-of-the-art binary neural networks in object detection.
\end{itemize}

\section{Related Work}
\textbf{Network Quantization: }Network quantization has been widely studied in recent years due to its efficiency in storage and computation. Existing methods can be divided into two categories: neural networks with weights and activations in one bit or multiple bits. Binary neural networks reduce the model complexity significantly due to the extremely high compression ratio. Hubara \etal \cite{hubara2016binarized} and Rastegari \etal \cite{rastegari2016xnor} binarized both weights and activations in neural networks and replaced the multiply-accumulation with xnor and bitcount operations, where straight-through estimators were applied to relax the non-differentiable sign function for back-propagation. Liu \etal \cite{liu2018bi} added extra shortcut between consecutive convolutional blocks to strengthen the representational capacity of the network. They also used custom gradients to optimize the non-differentiable networks. Binary neural networks perform poorly on difficult tasks such as object detection due to the low representational capacity, multi-bit quantization strategies have been proposed with wider bitwidth. Jacob \etal \cite{jacob2018quantization} presented an 8-bit quantized model for inference in object detection and their method can be integrated with efficient architectures. Wei \etal \cite{wei2018quantization} applied the knowledge distillation to learn 8-bit neural networks in small size from large full-precision models. Li \etal \cite{li2019fully} proposed fully quantized neural networks in four bits with hardware-friendly implementation. Meanwhile, the instabilities during training were overcome by the presented techniques. Nevertheless, multi-bit neural networks still suffer from heavy storage and computation cost. Directly applying binary neural networks with constrained representational power in object detection leads to numerous false positives and significantly degrades the performance due to the information redundancy in the networks.

\textbf{Object Detection: }Object detection has aroused comprehensive interest in computer vision due to its wide application. Modern CNN based detectors are categorized into two-stage and one-stage detectors. In the former, R-CNN \cite{girshick2014rich} was among the earliest CNN-based detectors with the pipeline of bounding box regression and classification. Progressive improvements were proposed for better efficiency and effectiveness. Fast R-CNN \cite{girshick2015fast} presented the ROIpooling in the detection framework to achieve better accuracy and faster inference. Faster R-CNN \cite{ren2015faster} proposed the Region Proposal Networks to effectively generate region proposals instead of hand-crafted ones. FPN \cite{lin2017feature} introduced top-down architectures with lateral connections and the multi-scale features to integrate low-level and high-level features. In the latter regard, SSD \cite{liu2016ssd} and YOLO \cite{redmon2016you} directly predicted the bounding box and the class without region proposal generation, so that real-time inference was achieved on GPUs with competitive accuracy. RetinaNet \cite{lin2017focal} proposed the focal loss to solve the problem of foreground-background class imbalance. However, CNN based detectors suffer from heavy storage and computational cost so that their deployment is limited.

\begin{figure*}[t]
	\centering
	\includegraphics[height=5cm, width=14.5cm]{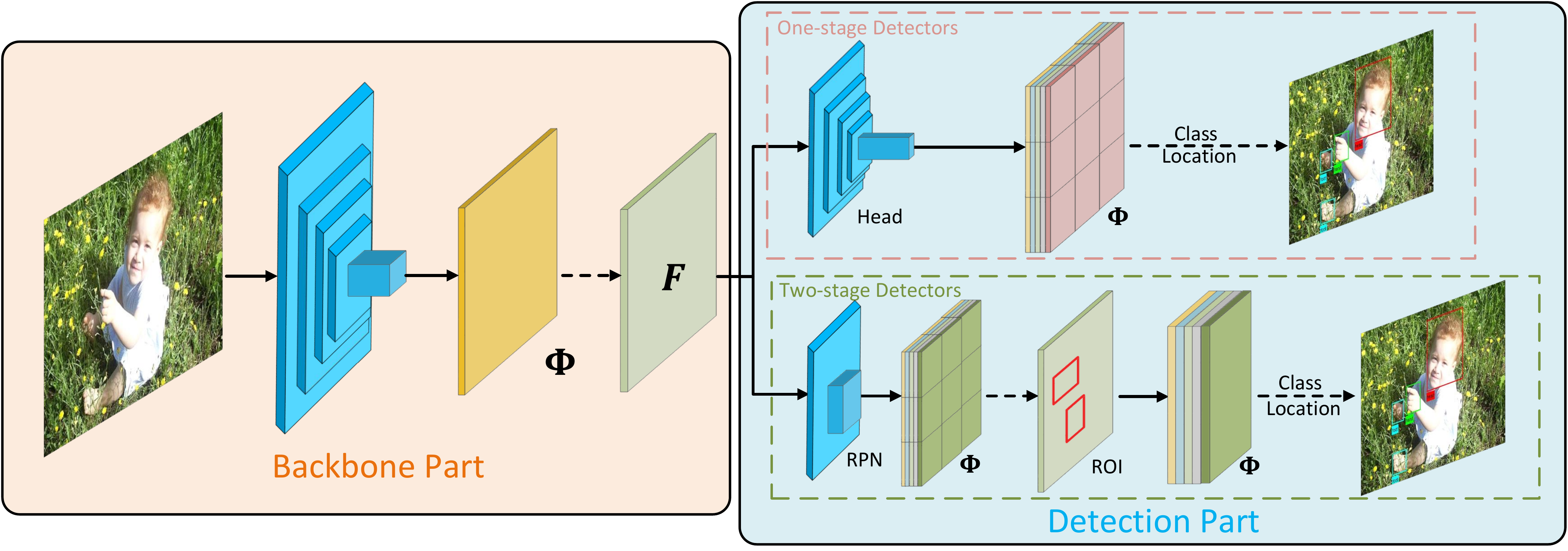}
	\caption{The pipeline of the information bottleneck based detectors, which consist of the backbone part and the detection part. The solid line represents the forward propagation in the network, while the dashed line means sampling from a parameterized distribution $\Phi$. The high-level feature map $F$ is sampled from the distribution parameterized by the backbone network. The one-stage and two-stage detector framework can be both employed in the detection part of our BiDet. For the one-stage detectors, the head network parameterizes the distribution of object classes and location. For two-stage detectors, Region Proposal Networks (RPN) parameterize the prior distribution of location and the posteriors are parameterized by the refining networks. (best viewed in color).}
	\vspace{-0.3cm}
	\label{pipeline}
\end{figure*}

\textbf{Information Bottleneck: }The information bottleneck (IB) principle was first proposed by \cite{tishby2000information} with the goal of extracting relevant information of the input with respect to the task, so that the IB principle are widely applied in compression. The IB principle enforces the mutual information between the input and learned features to be minimized while simultaneously maximizing the mutual information between the features and groundtruth of the tasks. Louizos \etal \cite{louizos2017bayesian} and Ullrich \etal \cite{ullrich2017soft} utilized the Minimal Description Length (MDL) principle that is equivalent to IB to stochastically quantize deep neural networks. Moreover, they used the sparse horseshoe and Gaussian mixture priors for weight learning in order to reduce the quantization errors. Dai \etal \cite{dai2018compressing} pruned individual neurons via variational IB so that redundancy between adjacent layers was minimized by aggregating useful information in a subset of neurons. Despite the network compression, IB is also utilized in compact feature learning. Amjad \etal \cite{amjad2019learning} proposed stochastic deep neural networks where IB could be utilized to learn efficient representations for classification. Shen \etal \cite{shen2019embarrassingly} imposed IB on existing hash models to generate effective binary representations so that the data semantics were fully utilized. In this paper, we extend the IB principle to squeeze the redundancy in binary detection networks, so that the false positives are alleviated and the detection precision is significantly enhanced.

\section{Approach}
In this section, we first extend the IB principle that removes the information redundancy to object detection. Then we present the details of learning the sparse object priors for object detection, which concentrate posteriors on informative prediction with false positive elimination. Finally, we propose the efficient binarized object detectors.
\subsection{Information Bottleneck for Object Detection}
The information bottleneck (IB) principle directly relates to compression with the best hypothesis that the data misfit and the model complexity should simultaneously be minimized, so that the redundant information irrelevant to the task is exclusive in the compressed model and the capacity of the lightweight model is fully utilized. The task of object detection can be regarded as a Markov process with the following Markov chain:
\begin{align}
	X\rightarrow F\rightarrow L,C
\end{align}
where $X$ means the input images and $F$ stands for the high-level feature maps output by the backbone part. $C$ and $L$ represent the predicted classes and location of the objects respectively. According to the Markov chain, the objective of the IB principle is written as follows:
\begin{align}
	\min\limits_{\phi_b,\phi_d} ~~ I(X;F)-\beta I(F;C,L)
	\label{IB}
\end{align} where $\phi_b$ and $\phi_d$ are the parameters of the backbone and the detection part respectively. $I(X;Y)$ means the mutual information between two random variables $X$ and $Y$. Minimizing the mutual information between the images and the high-level feature maps constrains the amount of information that the detector extracts, and maximizing the mutual information between the high-level feature maps and object detection enforces the detector to preserve more information related to the task. As a result, the redundant information irrelevant to object detection is removed. Figure \ref{pipeline} shows the pipeline for information bottleneck based detectors, the IB principle can be imposed on the conventional one-stage and two-stage detectors. We rewrite the first term of (\ref{IB}) according to the definition of mutual information:
\begin{align}
	I(X;F)=\mathbb{E}_{\bm{x}\sim p(\bm{x})}\mathbb{E}_{\bm{f}\sim p(\bm{f}|\bm{x})}\log\frac{ p(\bm{f}|\bm{x})}{p(\bm{f})}
	\label{feature_map}
\end{align}where $\bm{x}$ and $\bm{f}$ are the specific input images and the corresponding high-level feature maps. $p(\bm{x})$ and $p(\bm{f})$ are the prior distribution of $\bm{x}$ and $\bm{f}$ respectively, and $\mathbb{E}$ represents the expectation.  $p(\bm{f}|\bm{x})$ is the posterior distribution of the high-level feature map conditioned on the input. We parameterize $p(\bm{f}|\bm{x})$ by the backbone due to its intractability, where evidence-lower-bound (ELBO) minimization is applied for relaxation. To estimate $I(X;F)$, we sample the training set to obtain the image $\bm{x}$ and sample the distribution parameterized by the backbone to acquire the corresponding high-level feature map $\bm{f}$.

 	 \begin{figure}[t]
		\centering
		\includegraphics[height=6.5cm, width=8cm]{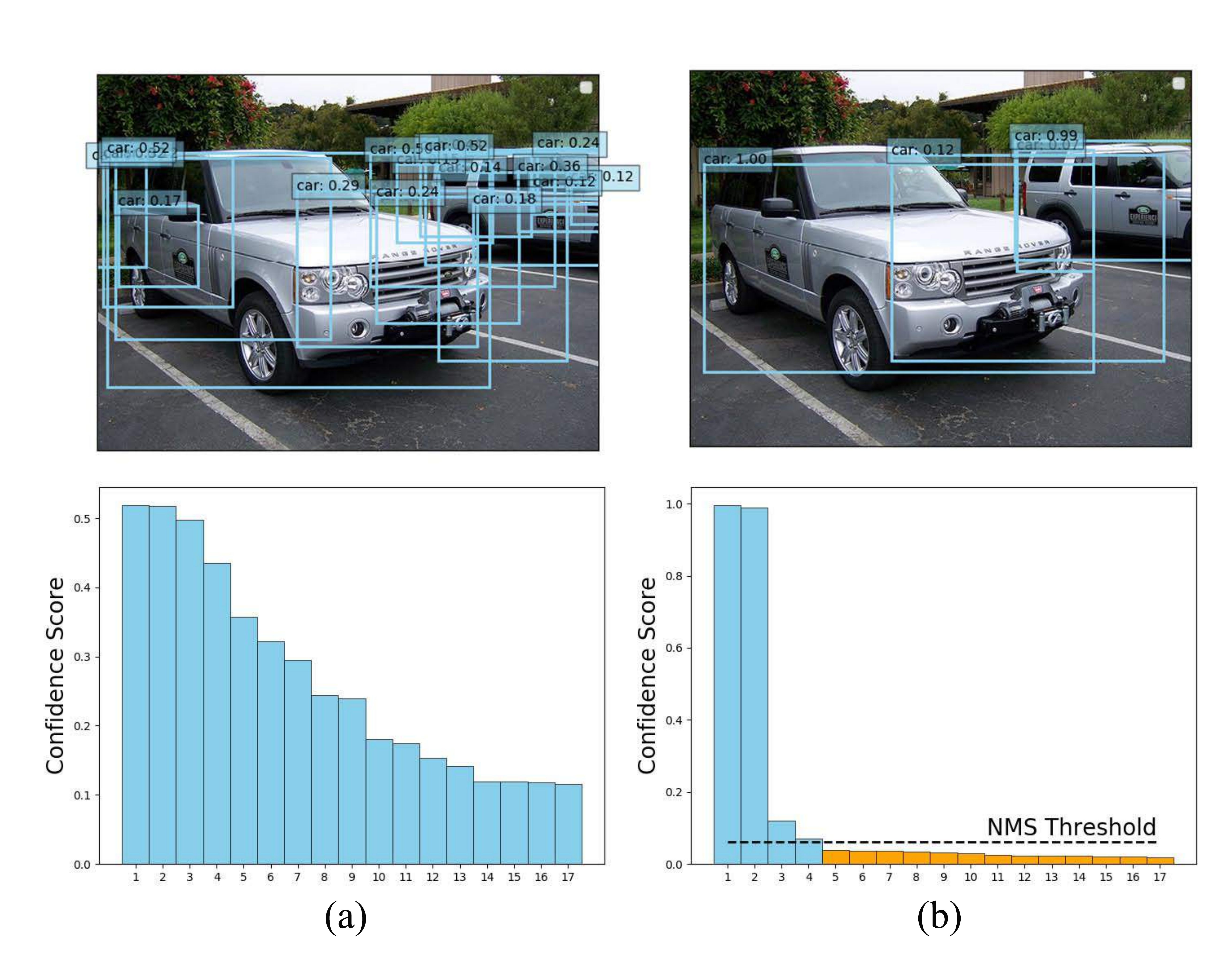}
	\caption{The detected objects and the corresponding confidence score (a) before and (b) after optimizing (\ref{alter_sparse}). The contrast of confidence score among different detected objects is significantly enlarged by minimizing alternate objective. As the NMS eliminates the positives with confidence score lower than the threshold, the sparse object priors are acquired and the posteriors are enforced to be concentrated on informative prediction. (best viewed in color).}
	\vspace{-0.3cm}
	\label{priors}
\end{figure}

The location and classification of objects based on the high-level feature map are independent, as the bounding box location and the classification probability are obtained via different network branches in the detection part. The mutual information in the second term of (\ref{IB}) is factorized:
\begin{align}
	I(F;C,L)=I(F;C)+I(F;L)
\end{align}Similar to (\ref{feature_map}), we rewrite the mutual information between the high-level feature maps and the classes as follows:
\begin{align}
	I(F;C)=\mathbb{E}_{\bm{f}\sim p(\bm{f}|\bm{x})}\mathbb{E}_{\bm{c}\sim p(\bm{c}|\bm{f})}\log\frac{p(\bm{c}|\bm{f})}{p(\bm{c})}
	\label{mi_class}
\end{align}where $\bm{c}$ is the object class labels including the background class. $p(\bm{c})$ and $p(\bm{c}|\bm{f})$ are the prior class distribution and posterior class distribution when given the feature maps respectively. Same as the calculation of (\ref{feature_map}), we employ the classification branch networks in the detection part to parameterize the distribution. Meanwhile, we divide the images to blocks for multiple object detection. For one-stage detectors such as SSD \cite{liu2016ssd}, we project the high-level feature map cells to the raw image to obtain the block partition. For two-stage detectors such as Faster R-CNN \cite{ren2015faster}, we scale the ROI to the original image for block split. $\bm{c}\in \mathbb{Z}^{1\times b}$ represents the object class in $b$ blocks of the image. We define $c_{i}$ as the $i_{th}$ element of $\bm{c}$, which demonstrates the class of the object whose center is in the $i_{th}$ block of the image. The class of a block is assigned to background if the block does not contain the center of any groundtruth objects.

As the localization contains shift parameters and scale parameters for anchors, we rewrite the mutual information between the object location and high-level feature maps:

\vspace{-0.3cm}
\begin{small}
\begin{align*}
	I(F;L)=\mathbb{E}_{\bm{f}\sim p(\bm{f}|\bm{x})}\mathbb{E}_{\bm{l}_1\sim p(\bm{l}_1|\bm{f})}\mathbb{E}_{\bm{l}_2\sim p(\bm{l}_2|\bm{f})}\log\frac{p(\bm{l}_1|\bm{f})p(\bm{l}_2|\bm{f})}{p(\bm{l}_1)p(\bm{l}_2)}
\end{align*}
\end{small}where $\bm{l}_1 \in \mathbb{R}^{2\times b}$ represents the horizontal and vertical shift offset of the anchors in $b$ blocks of the image, and $\bm{l}_2 \in \mathbb{R}^{2\times b}$ means the height and width scale offset of the anchors. For the anchor whose center  $(x,y)$ is in the $j_{th}$ block with height $h$ and width $w$, the offset changes the bounding box in the following way: $(x,y)\rightarrow (x,y)+\bm{l}_{1,j}$ and $(h,w)\rightarrow (h,w)\cdot exp(\bm{l}_{2,j})$, where $\bm{l}_{1,j}$ and $\bm{l}_{2,j}$ represent the $j_{th}$ column of $\bm{l}_1$ and $\bm{l}_2$. The priors and the posteriors of shift offset conditioned on the feature maps  are denoted as $p(\bm{l}_1)$ and $p(\bm{l}_1|\bm{f})$ respectively. Similarly, the scaling offset has the prior and the posteriors given feature maps $p(\bm{l}_2)$ and $p(\bm{l}_2|\bm{f})$.
We leverage the localization branch networks in the detection part for distribution parameterization.

\subsection{Learning Sparse Object Priors}
Since the feature maps are binarized in BiDet, we utilize the binomial distribution with equal probability as the priors for each element of the high-level feature map $\bm{f}$. We assign the priors for object localization in the following form: $p(\bm{l}_{1,j})= N(\bm{\mu}_{1,j}^0,\bm{\Sigma}_{1,j}^0)$ and $p(\bm{l}_{2,j})= N(\bm{\mu}_{2,j}^0,\bm{\Sigma}_{2,j}^0)$, where $N(\bm{\mu}, \bm{\Sigma})$ means the Gaussian distribution with mean $\bm{\mu}$ and covariance matrix $\bm{\Sigma}$. For one-stage detectors, the object localization priors $p(\bm{l}_{1,j})$ and $p(\bm{l}_{2,j})$ are hypothesized to be the two-dimensional standard normal distribution. For two-stage detectors, Region Proposal Networks (RPN) output the parameters of the Gaussian priors.

As numerous false positives emerge in the binary detection networks, learning sparse object priors for detection part enforces the posteriors to be concentrated on informative detection prediction with false positive elimination. The priors for object classification is defined as follows:
\begin{align*}
	p(c_i)=\mathbb{I}_{M_i}\cdot cat(\frac{1}{n+1}\cdot\bm{1}^{n+1})+(1-\mathbb{I}_{M_i})\cdot cat([1,\bm{0}^{n}])
\end{align*}where $\mathbb{I}_x$ is the indicator function with $\mathbb{I}_1=1$ and $\mathbb{I}_0=0$,  and $M_i$ is the $i_{th}$ element of the block mask $\bm{M}\in\{0,1\}^{1\times b}$. $cat(\bm{K})$ means the categorical distribution with the parameter $\bm{K}$. $\bm{1}^{n}$ and $\bm{0}^{n}$ are the all-one and zero vectors in $n$ dimensions respectively, where $n$ is the number of class. The multinomial distribution with equal probability is utilized for the class prior in the $i_{th}$ block if $M_i$ equals to one.  Otherwise, the categorical distribution with the probability $1$ for background and zero probability for other classes is leveraged for the prior class distribution. When $M_i$ equals to zero, the detection part definitely predicts the background for object classification in the $i_{th}$ block according to (\ref{mi_class}). In order to obtain sparse priors for object classification with fewer predicted positives, we minimize the $L_1$ norm of the block mask $\bm{M}$. We propose an alternative way to optimize $\bm{M}$ due to the non-differentiability, where the objective is written as follows:
\begin{align}
	\min\limits_{s_i} -\frac{1}{m}\sum_{i=1}^{m}s_i \log s_i
	\label{alter_sparse}
\end{align}where $m=||\bm{M}||_1$ represents the number of detected foreground objects in the image, and $s_i$ is the normalized confidence score for the $i_{th}$ predicted foreground object with $\sum_{i=1}^{m}s_i=1$. As shown in Figure \ref{priors}, minimizing (\ref{alter_sparse}) increases the contrast of confidence score among different predicted objects, and predicted objects with low confidence score are assigned to be negative by the non-maximum suppression (NMS) algorithms. Therefore, the block mask becomes sparser with fewer predicted objects, and the posteriors are concentrated on informative prediction with uninformative false positive elimination.

\subsection{Efficient Binarized Object Detectors}
In this section, we first briefly introduce neural networks with binary weights and activations, and then detail the learning objectives of our BiDet.  Let $\bm{W^l_r}$ be the real-valued weights and $\bm{A^l_r}$ be the full-precision activations of the $l_{th}$ layer in a given L-layer detection model. During the forward propagation, the weights and activations are binarized via the sign function: $\bm{W^l_b}=sign(\bm{W^l_r})$ and $\bm{A^l_b}=sign(\bm{W^l_r}\odot\bm{A^l_b})$. $sign$ means the element-wise sign function which maps the number larger than zero to one and otherwise to minus one, and $\odot$ indicates the element-wise binary product consisting of xnor and bitcount operations. Due to the non-differentiability of the sign function, straight-through estimator (STE) is employed to calculate the approximate gradients and update the real-valued weights in the back-propagation stage. The learning objectives for the proposed BiDet is written as follows:
\vspace{-0.5cm}

\begin{small}
\begin{align}
&\min J = J_1+J_2\notag\\
&=(\sum_{t,s}\log \frac{ p(f_{st}|\bm{x})}{p(f_{st})}-\beta\sum_{i=1}^{b}\log\frac{p(c_i|\bm{f})p(\bm{l}_{1,i}|\bm{f})p(\bm{l}_{2,i}|\bm{f})}{p(c_i)p(\bm{l}_{1,i})p(\bm{l}_{2,i})})\notag\\
& \quad-\gamma\cdot\frac{1}{m}\sum_{i=1}^{m}s_i \log s_i
\label{objective}
\end{align}
\end{small}where $\gamma$ is a hyperparameter that balances the importance of false positive elimination. The posterior distribution $p(c_i|\bm{f})$ is hypothesized to be the categorical distribution $cat(\bm{K}_i)$, where $\bm{K}_i\in \mathbb{R}^{1\times(n+1)}$ is the parameter and $n$ is the number of classes. We assume the posterior of the shift and scale offset follows the Gaussian distribution: $p(\bm{l}_{1,j}|\bm{f})= N(\bm{\mu}_{1,j},\bm{\Sigma}_{1,j})$ and $p(\bm{l}_{2,j}|\bm{f})= N(\bm{\mu}_{2,j},\bm{\Sigma}_{2,j})$. The posteriors of the element in the $s_{th}$ row and $t_{th}$ column of binary high-level feature maps $p(f_{st}|\bm{x})$ is assigned to binomial distribution $cat([p_{ts},1-p_{ts}])$, where $p_{ts}$ is the probability for $f_{st}$ to be one. All the posterior distribution is parameterized by the neural networks. $J_1$ represents for the information bottleneck employed in object detection, which aims to remove information redundancy and fully utilize the representational power of the binary neural networks. The goal of $J_2$ is to enforce the object priors to be sparse so that the posteriors are encouraged to be concentrated on informative prediction with false positive elimination.

In the learning objective, $p(f_{st})$ in the binomial distribution is a constant. Meanwhile, the sparse object classification priors are imposed via $J_2$ so that $p(c_i)$ is also regarded as a constant. For one-stage detectors, constant $p(\bm{l}_{1,i})$ and $p(\bm{l}_{2,i})$ follows standard normal distribution. For two-stage detectors, $p(\bm{l}_{1,i})$ and $p(\bm{l}_{2,i})$ are parameterized by RPN, which is learned by the objective function. The last layer of the backbone that outputs the parameters of the binary high-level feature maps is real-valued in training for Monte-Carlo sampling and is binarzed with the sign function during inference. Meanwhile, the layers that output the parameters for object class and location distribution remain real-valued for accurate detection. During inference, we drop the network branch of covariance matrix for location offset, and assign all location prediction with the mean value to accelerate computation. Moreover, the prediction of object classes is set to that with the maximum probability to avoid time-consuming stochastic sampling in inference.

\section{Experiments}
In this section, we conducted comprehensive experiments to evaluate our proposed method on two datasets for object detection: PASCAL VOC \cite{everingham2010pascal} and COCO \cite{lin2014microsoft}. We first describe the implementation details of our BiDet, and then we validate the effectiveness of IB and sparse object priors for binarized object detectors by ablation study. Finally, we compare our method with state-of-the-art binary neural networks in the task of object detection to demonstrate superiority of the proposed BiDet.

\subsection{Datasets and Implementation Details}
We first introduce the datasets that we carried out experiments on and data preprocessing techniques:

\textbf{PASCAL VOC: }The PASCAL VOC dataset contains natural images from 20 different classes. We trained our model on the VOC 2007 and VOC 2012 trainval sets which consist of around 16k images, and we evaluated our method on VOC 2007 test set including about 5k images. Following \cite{everingham2010pascal}, we used the mean average precision (mAP) as the evaluation criterion.

\textbf{COCO: }The COCO dataset consists of images from 80 different categories. We conducted experiments on the 2014 COCO object detection track. We trained our model with the combination of 80k images from the training set and 35k images sampled from validation set (trainval35k \cite{bell2016inside}) and tested our method on the remaining 5k images in the validation set (minival \cite{bell2016inside}). Following the standard COCO evaluation metric \cite{lin2014microsoft},  we report the average precision (AP) for IoU $\in  \left[0.5 : 0.05 : 0.95\right]$ denoted as mAP@$\left[.5, .95\right]$. We also report AP$_{50}$, AP$_{75}$ as well as AP$_{s}$, AP$_{m}$ and AP$_{l}$ to further analyze our method.

We trained our BiDet with the SSD300 \cite{liu2016ssd} and Faster R-CNN \cite{ren2015faster} detection framework whose backbone were VGG16 \cite{simonyan2014very} and ResNet-18 \cite{he2016deep} respectively. Following the implementation of binary neural networks in \cite{hubara2016binarized}, we remained the first and last layer in the detection networks real-valued. We used the data augmentation techniques in \cite{liu2016ssd} and \cite{ren2015faster} when training our BiDet with SSD300 and Faster R-CNN detection frameworks respectively.

In most cases, the backbone network was pre-trained on ImageNet \cite{russakovsky2015imagenet} in the task of image classification. Then we jointly finetuned the backbone part and trained the detection part for the object detection task. The batchsize was assigned to be $32$, and the Adam optimizer \cite{kingma2014adam} was applied. The learning rate started from 0.001 and decayed twice by multiplying $0.1$ at the $6_{th}$ and $10_{th}$ epoch out of $12$ epochs. Hyperparamters $\beta$ and $\gamma$ were set as $10$ and $0.2$ respectively.

 	 \begin{figure}[t]
		\centering
		\includegraphics[height=7.5cm, width=8.5cm]{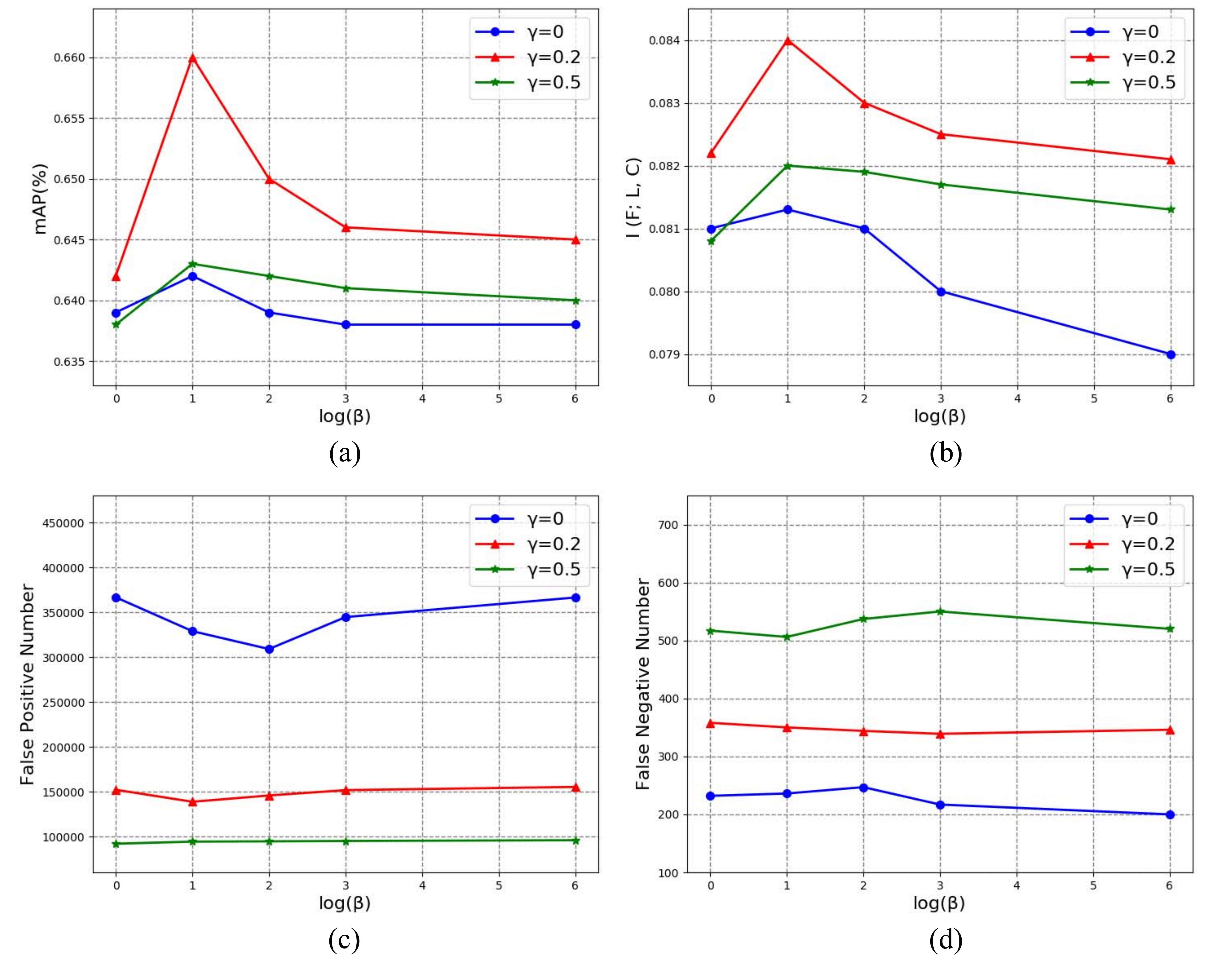}
	\caption{Ablation study w.r.t. hyperparameters $\beta$ and $\gamma$, where the variety of (a) mAP, (b) the mutual information between high-level feature maps and the object detection $I(F;L,C)$ , (c) the number of false positives and (d) the number of false negatives are demonstrated. (best viewed in color).}
	\vspace{-0.3cm}
	\label{ablation}
\end{figure}

\begin{table*}[t]
	\footnotesize
	\caption{Comparison of parameter size, FLOPs and mAP (\%) with the state-of-the-art binary neural networks in both one-stage and two-stage detection frameworks on PASCAL VOC. The detector with the real-valued and multi-bit backbone is given for reference. BiDet (SC) means the proposed method with extra shortcut for the architectures. }
	\label{MAPVOC}
	\centering
	\vspace{0.1cm}
\renewcommand\arraystretch{1.2}
\begin{tabular}{|c|c|c|c|c|c|c|c|}
	\hline
	Framework & Input & Backbone & Quantization & W/A (bit) & \#Params & MFLOPs & mAP\\
	\hline
	\multirow{10}{*}{SSD300} & \multirow{10}{*}{$300\times300$} & VGG16 & \multirow{2}{*}{$-$} & \multirow{2}{*}{$32/32$} & $100.28$MB & $31,750$ & $72.4$\\
	& & MobileNetV1 & & & $30.07$MB & $1,150$ & $68.0$\\
	\cline{3-8}
	& & \multirow{7}{*}{VGG16} & TWN & $2/32$ & $24.54$MB & $8,531$ & $67.8$\\
	& & & DoReFa-Net & $4/4$ & $29.58$MB & $4,661$ & $69.2$\\
	\cline{4-8}
	& & & BNN & \multirow{3}{*}{$1/1$} & $22.06$MB & $1,275$ & $42.0$\\
	& & & Xnor-Net & & $22.16$MB & $1,279$ & $50.2$\\
	& & & BiDet & & $22.06$MB & $1,275$ & $\bm{52.4}$\\
	\cline{4-8}
	& & & Bi-Real-Net & \multirow{2}{*}{$1/1$}  & $21.88$MB & $1,277$ & $63.8$\\
	
	& & & BiDet (SC) & & $21.88$MB & $1,277$ & $\bm{66.0}$\\
	\cline{3-8}
	& & \multirow{2}{*}{MobileNetV1} & Xnor-Net & \multirow{2}{*}{$1/1$}  & $22.48$MB & $836$ & $48.9$\\
	& &  & BiDet & & $22.48$MB & $836$ & $\bm{51.2}$\\
	\hline
	\multirow{8}{*}{Faster R-CNN} & \multirow{8}{*}{$600\times1000$} & \multirow{8}{*}{ResNet-18} & $-$ & $32/32$ & $47.35$MB & $36,013$ & $74.5$\\
	\cline{4-8}
	& & & TWN & $2/32$ & $3.83$MB & $9,196$ & $69.9$\\
	& & & DoReFa-Net & $4/4$ & $6.73$MB & $4,694$ & $71.0$\\
	\cline{4-8}
	& & & BNN & \multirow{3}{*}{$1/1$} & $2.38$MB & $779$ & $35.6$\\
	& & & Xnor-Net & & $2.48$MB & $783$ & $48.4$\\
	& & & BiDet  & & $2.38$MB & $779$ & $\bm{50.0}$\\
	\cline{4-8}
	& & & Bi-Real-Net & \multirow{2}{*}{$1/1$} & $2.39$MB & $781$ & $58.2$\\
	& & & BiDet (SC) & & $2.39$MB & $781$ & $\bm{59.5}$\\
	\hline
\end{tabular}
	\vspace{-0.2cm}
\end{table*}

\begin{table*}[t]
	\footnotesize
	\caption{Comparison of mAP@$\left[.5, .95\right]$ (\%), AP with different IOU threshold and AP for objects in various sizes with state-of-the-art binarized object detectors in SSD300 and Faster R-CNN detection framework on COCO, where the performance of real-valued and multi-bit detectors is reported for reference. BiDet (SC) means the proposed method with extra shortcut for the architectures.}
	\label{MAPCOCO}
	\centering
	\vspace{0.1cm}
	\renewcommand\arraystretch{1.2}
\begin{tabular}{|c|c|c|c|c|cc|ccc|}
	\hline
	Framework & Input & Backbone & Quantization & mAP@$\left[.5, .95\right]$ & AP$_{50}$ & AP$_{75}$ (\%) & AP$_{s}$ & AP$_{m}$ & AP$_{l}$\\
	\hline
	\multirow{8}{*}{SSD300} & \multirow{8}{*}{$300\times300$} & \multirow{8}{*}{VGG16} & $-$ & $23.2$ & $41.2$ & $23.4$ & $5.3$ & $23.2$ & $39.6$\\
	\cline{4-10}
	& &  & TWN & $16.9$ & $33.0$ & $15.8$ & $5.0$ & $16.9$ & $27.2$\\
	& & & DoReFa-Net & $19.5$ & $35.0$ & $19.6$ & $5.1$ & $20.5$ & $32.8$\\
	\cline{4-10}
	& &  & BNN & $6.2$ & $15.9$ & $3.8$ & $2.4$ & $10.0$ & $9.9$\\
	& & & Xnor-Net & $8.1$ & $19.5$ & $5.6$ & $2.6$ & $8.3$ & $13.3$\\
	& & & BiDet  & $\bm{9.8}$ & $\bm{22.5}$ & $\bm{7.2}$ & $\bm{3.1}$ & $\bm{10.8}$ & $\bm{16.1}$\\
	\cline{4-10}
	& & & Bi-Real-Net & $11.2$ & $26.0$ & $8.3$ & $3.1$ & $12.0$ & $18.3$\\
	& & & BiDet (SC)& $\bm{13.2}$ & $\bm{28.3}$ & $\bm{10.5}$ & $\bm{5.1}$ & $\bm{14.3}$ & $\bm{20.5}$\\
	\hline
	\multirow{8}{*}{Faster R-CNN} & \multirow{8}{*}{$600\times1000$} & \multirow{8}{*}{ResNet-18} & $-$ & $26.0$ & $44.8$ & $27.2$ & $10.0$ & $28.9$ & $39.7$\\
	\cline{4-10}
	& & & TWN & $19.7$ & $35.3$ & $19.7$ & $5.1$ & $20.7$ & $33.3$\\
	& & & DoReFa-Net & $22.9$ & $38.6$ & $23.7$ & $8.0$ & $24.9$ & $36.3$\\
	\cline{4-10}
	& & & BNN & $5.6$ & $14.3$ & $2.6$ & $2.0$ & $8.5$ & $9.3$\\
	& & & Xnor-Net & $10.4$ & $21.6$ & $8.8$ & $2.7$ & $11.8$ & $15.9$\\
	& & & BiDet & $\bm{12.1}$ & $\bm{24.8}$ & $\bm{10.1}$ & $\bm{4.1}$ & $\bm{13.5}$ & $\bm{17.7}$\\
	\cline{4-10}
	& & & Bi-Real-Net & $14.4$ & $29.0$ & $13.4$ & $3.7$ & $15.4$ & $24.1$\\
	
	& & & BiDet (SC) & $\bm{15.7}$ & $\bm{31.0}$ & $\bm{14.4}$ & $\bm{4.9}$ & $\bm{16.7}$ & $\bm{25.4}$\\
	\hline
\end{tabular}
	\vspace{-0.2cm}
\end{table*}

\subsection{Ablation Study}
Since the IB principle removes the redundant information in binarized object detectors and the learned sparse object priors concentrate the posteriors on informative prediction with false positive alleviation, the detection accuracy is enhanced significantly. To verify the effectiveness of the IB principle and the learned sparse priors, we conducted the ablation study to evaluate our BiDet w.r.t. the hyperparameter $\beta$ and $\gamma$ in the objective function. We adopted the SSD detection framework with VGG16 backbone for our BiDet on the PASCAL VOC dataset. We report the mAP, the mutual information between high-level feature maps and the object detection $I(F;L,C)$, the number of false positives and the number of false negatives with respect to $\beta$ and $\gamma$ in Figure \ref{ablation} (a), (b), (c) and (d) respectively. Based on the results, we observe the influence of the IB principle and the learned sparse object priors as follows.

By observing Figure \ref{ablation} (a) and (b), we conclude that mAP and $I(F;L,C)$ are positively correlated as they demonstrate the detection performance and the amount of related information respectively. Medium $\beta$ provides the optimal trade-off between the amount of extracted information and the related information so that the representational capacity of the binary object detectors is fully utilized with redundancy removal. Small $\beta$ fails to leverage the representational power of the networks as the amount of extracted information is limited by regularizing the high-level feature maps, while large $\beta$ enforces the networks to learn redundant information which leads to significant over-fitting. Meanwhile, medium $\gamma$ offers optimal sparse object priors that enforces the posteriors to concentrate on most informative prediction. Small $\gamma$ is not capable of sparsifying the predicted objects, and large $\gamma$ disables the posteriors to represent informative objects with excessive sparsity.

By comparing the variety of false positives and false negatives w.r.t. $\beta$ and $\gamma$, we know that medium $\beta$ decreases false positives most significantly and changing $\beta$ does not varies the number of false negatives notably, which means that the redundancy removal only alleviates the uninformative false positives while remains the informative true positives unchanged. Meanwhile, small $\gamma$ fails to constrain the false positives and large $\gamma$ clearly increases the false negatives, which both degrades the performance significantly.

\begin{figure*}[t]
	\centering
	\includegraphics[height=5cm, width=17.5cm]{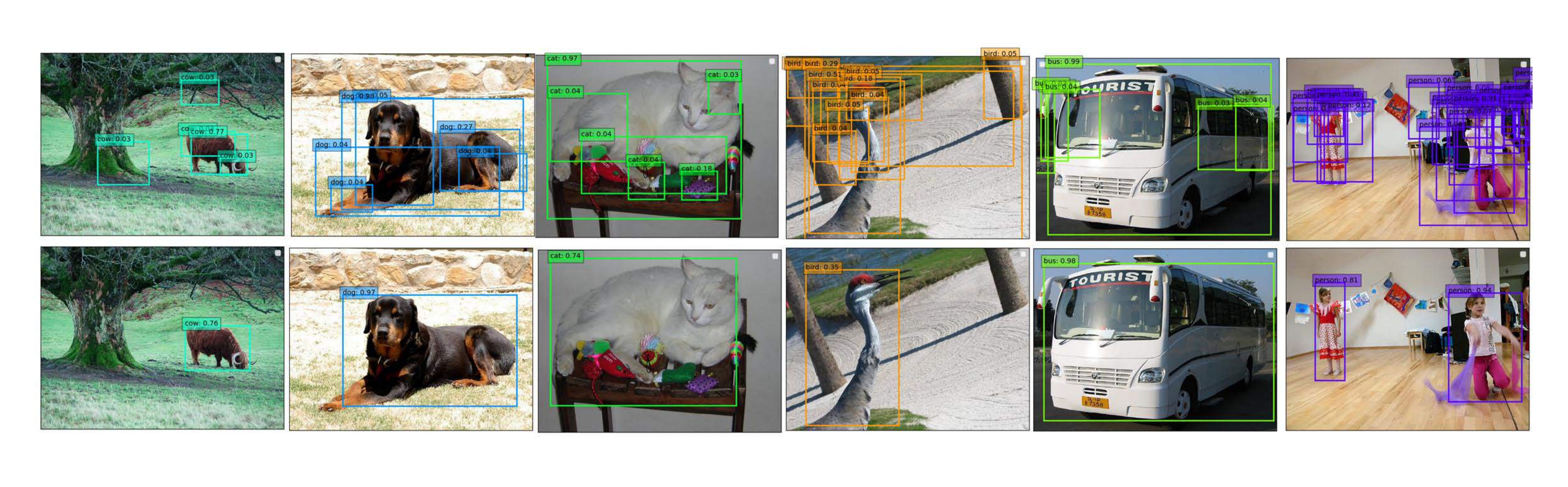}
	\caption{Qualitative results on PASCAL VOC. Images in the top row shows the object predicted by Xnor-Net, while the images with the objects detected by our BiDet are displayed in the bottom row. The proposed BiDet removes the information redundancy to fully utilize the network capacity, and learns the sparse object priors to eliminate false positives (best viewed in color).}
	\label{visualization}
\end{figure*}

\subsection{Comparison with the State-of-the-art Methods}
In this section, we compare the proposed BiDet with the state-of-the-art binary neural networks including BNN \cite{courbariaux2015binaryconnect}, Xnor-Net \cite{rastegari2016xnor} and Bi-Real-Net \cite{liu2018bi} in the task of object detection on the PASCAL VOC and COCO datasets. For reference, we report the detection performance of the multi-bit quantized networks containing DoReFa-Net \cite{zhou2016dorefa} and TWN \cite{li2016ternary} and the lightweight networks MobileNetV1 \cite{howard2017mobilenets}.

\textbf{Results on PASCAL VOC: }Table \ref{MAPVOC} illustrates the comparison of computation complexity, storage cost and the mAP across different quantization methods and detection frameworks.  Our BiDet significantly accelerates the computation and saves the storage by $24.90\times$ and $4.55\times$ with the SSD300 detector and $46.23\times$ and $19.81\times$ with the Faster R-CNN detector. The efficiency is enhanced more notably in the Faster R-CNN detector, as there are multiple real-valued output layers of the head networks in SSD300 for multi-scale feature extraction.

Compared with the state-of-the-art binary neural networks, the proposed BiDet improves the mAP of Xnor-Net by $2.2\%$ and $1.6\%$ with SSD300 and Faster R-CNN frameworks respectively with fewer FLOPs and the number of parameters than Xnor-Net. As demonstrated in \cite{liu2018bi}, adding extra shortcut between consecutive convolutional layers can further enhance the representational power of the binary neural networks, we also employ architecture with additional skip connection to evaluate our BiDet in networks with stronger capacity. Due to the information redundancy, the performance of Bi-Real-Net with constrained network capacity is degraded significantly compared with their full-precision counterparts in both one-stage and two-stage detection frameworks. On the contrary, our BiDet imposes the IB principle on learning binary neural networks for object detection and fully utilizes the network capacity with redundancy removal. As a result, the proposed BiDet increases the mAP of Bi-Real-Net by $2.2\%$ and $1.3\%$ in SSD300 and Faster R-CNN detectors respectively without additional computational and storage cost. Figure \ref{visualization} shows the qualitative results of Xnor-Net and our BiDet in the SSD300 detection framework with VGG16, where the proposed BiDet significantly alleviates the false positives.

Due to the different pipelines in one-stage and two-stage detectors, the mAP gained from the proposed BiDet with Faster R-CNN is less than SSD300. As analyzed in \cite{lin2017focal}, one-stage detectors face the severe positive-negative class imbalance problem which two-stage detectors are free of, so that one-stage detectors are usually more vulnerable to false positives. Therefore, one-stage object detection framework obtains more benefits from the proposed BiDet, which learns the sparse object priors to concentrate the posteriors on informative prediction with false positive elimination.

Moreover, our BiDet can be integrated with other efficient networks in object detection for further computation speedup and storage saving. We employ our BiDet as a plug-and-play module in SSD detector with the MobileNetV1 network and saves the computational and storage cost by $1.47\times$ and $1.38\times$ respectively. Compared with the detectors that directly binarize weights and activations in MobileNetV1 with Xnor-Net, BiDet improves the mAP by a sizable margin, which depicts the effectiveness of redundancy removal for networks with extremely low capacity.

\textbf{Results on COCO: }The COCO dataset is much more challenging for object detection than PASCAL VOC due to the high diversity and large scale. Table \ref{MAPCOCO} demonstrates mAP, AP with different IOU threshold and AP of objects in various sizes. Compared with the state-of-the-art binary neural networks Xnor-Net, our BiDet improves the mAP by $1.7\%$ and $1.7\%$ in SSD300 and Faster R-CNN detection framework respectively due to the information redundancy removal. Moreover, the proposed BiDet also enhances the binary one-stage and two-stage detectors with extra shortcut by $2.0\%$ and $1.3\%$ on mAP. Comparing with the baseline methods of network quantization, our method achieves better performance in the AP with different IOU threshold and AP for objects in different sizes, which demonstrates the universality in different application settings.

\section{Conclusion}
In this paper, we have proposed a binarized neural network learning method called BiDet for efficient object detection. The presented BiDet removes the redundant information via information bottleneck principle to fully utilize the representational capacity of the networks and enforces the posteriors to be concentrated on informative prediction for false positive elimination, through which the detection precision is significantly enhanced. Extensive experiments depict the superiority of BiDet in object detection compared with the state-of-the-art binary neural networks.

\section*{Acknowledgement}
This work was supported in part by the National Key Research and Development Program of China under Grant 2017YFA0700802, in part by the National Natural Science Foundation of China under Grant 61822603, Grant U1813218, Grant U1713214, and Grant 61672306, in part by the Shenzhen Fundamental Research Fund (Subject Arrangement) under Grant JCYJ20170412170602564, and in part by Tsinghua University Initiative Scientific Research Program.

{\small
\bibliographystyle{ieee_fullname}
\bibliography{egbib}
}

\end{document}